\documentclass[OJCS, a4paper]{IEEEoj}
\usepackage{cite}
\usepackage{amsmath,amssymb,amsfonts}
\usepackage{algorithmic}
\usepackage{graphicx,color}
\usepackage{textcomp}
\def\BibTeX{{\rm B\kern-.05em{\sc i\kern-.025em b}\kern-.08em
    T\kern-.1667em\lower.7ex\hbox{E}\kern-.125emX}}
\def\OJlogo{\vspace{-14pt}\includegraphics[height=28pt]{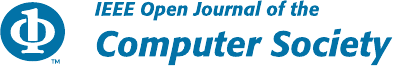}}

\usepackage[colorlinks=true,linkcolor=black,citecolor=blue,urlcolor=blue,]{hyperref}

\usepackage{multirow}
\usepackage{diagbox}

\begin{document}
\receiveddate{XX Month, XXXX}
\reviseddate{XX Month, XXXX}
\accepteddate{XX Month, XXXX}
\publisheddate{XX Month, XXXX}
\currentdate{XX Month, XXXX}
%\doiinfo{OJIM.2022.1234567}

\title{An Efficient Imbalance-Aware Federated Learning Approach for Wearable Healthcare with Autoregressive Ratio Observation}

\author{WENHAO YAN\authorrefmark{1}, MEMBER, IEEE, HE LI\authorrefmark{1}, MEMBER, IEEE, KAORU OTA\authorrefmark{1}, MEMBER, IEEE, AND MIANXIONG DONG\authorrefmark{1}, MEMBER, IEEE}

\affil{Department of Sciences and Informatics, Muroran Institute of Technology, Muroran 050-8585, Japan}

\corresp{CORRESPONDING AUTHOR: He Li (e-mail: heli@muroran-it.ac.jp).}

\authornote{This work is partially supported by JSPS KAKENHI Grant Numbers JP19K20250, JP20H04174, JP20K11784, JP22K11989, JP23K11063, Leading Initiative for Excellent Young Researchers (LEADER), MEXT, and JST, PRESTO Grant Number JPMJPR21P3, Japan.}

\markboth{An Efficient Imbalance-Aware Federated Learning Approach for Wearable Healthcare with Autoregressive Ratio Observation}{YAN \textit{et al.}}

\begin{abstract}
Widely available healthcare services are now getting popular because of advancements in wearable sensing techniques and mobile edge computing. People's health information is collected by edge devices such as smartphones and wearable bands for further analysis on servers, then send back suggestions and alerts for abnormal conditions. The recent emergence of federated learning allows users to train private data on local devices while updating models collaboratively. However, the heterogeneous distribution of the health condition data may lead to significant risks to model performance due to class imbalance. Meanwhile, as FL training is powered by sharing gradients only with the server, training data is almost inaccessible. The conventional solutions to class imbalance do not work for federated learning. In this work, we propose a new federated learning framework FedImT, dedicated to addressing the challenges of class imbalance in federated learning scenarios. FedImT contains an online scheme that can estimate the data composition during each round of aggregation, then introduces a self-attenuating iterative equivalent to track variations of multiple estimations and promptly tweak the balance of the loss computing for minority classes. Experiments demonstrate the effectiveness of FedImT in solving the imbalance problem without extra energy consumption and avoiding privacy risks. 

\end{abstract}


% \IEEEspecialpapernotice{(Invited Paper)}

\maketitle

\section{INTRODUCTION}
\IEEEPARstart{W}{ith} the wide adoption of the Internet of Things (IoT) and mobile edge computing (MEC), wearable devices have become increasingly popular in healthcare for common citizens \cite{b01} \cite{b04}. Wearable devices allowed more personal and efficient healthcare services for cheap manufacturing costs and designed privacy and security features such as data encryption or access controls \cite{b05} \cite{b06}. Moreover, we can continuously obtain the body records of participants using edge devices (e.g. smartphones or wearables) equipped with various sensors and analyze data in real time \cite{b02}. It has brought opportunities to enable emergency detection and assistance for abnormal health conditions. Early and immediate diagnosis of abnormal body conditions is essential for following medical treatments and saving lives in even more extreme situations \cite{b12}.

In order to accurately identify abnormal symptoms of users, machine learning methods were introduced into healthcare. Machine learning models rely on a large amount of natural sensing data from healthcare participants for training \cite{b14}. However, such demand for training data is often one of the main barriers to the adoption of intelligent healthcare. First, training in natural health data may lead to serious privacy concerns and legal risks for the sensitivity of personally identifiable information. In terms of existing data protection and privacy laws, the \emph{2016 General Data Protection Regulation (GDPR)} from the European Union (EU) considered personal health data as a special category of sensitive data requiring additional protection \cite{b10}. Healthcare providers were forced to ensure that they have a lawful basis for processing such data while limiting data sharing. What's more, data from diverse sources are stored in silos in the current healthcare system. Those data with different formats and locations pose challenges for organizations that are looking to use machine learning to gain insights without sharing available training data \cite{b03}. Lastly, considering the limited performance and battery of wearable edge devices, the overhead of transferring the raw data obtained from the sensors to the central server is a non-negligible cost.

\begin{figure}[!t]
\centering{\includegraphics[width=\linewidth]{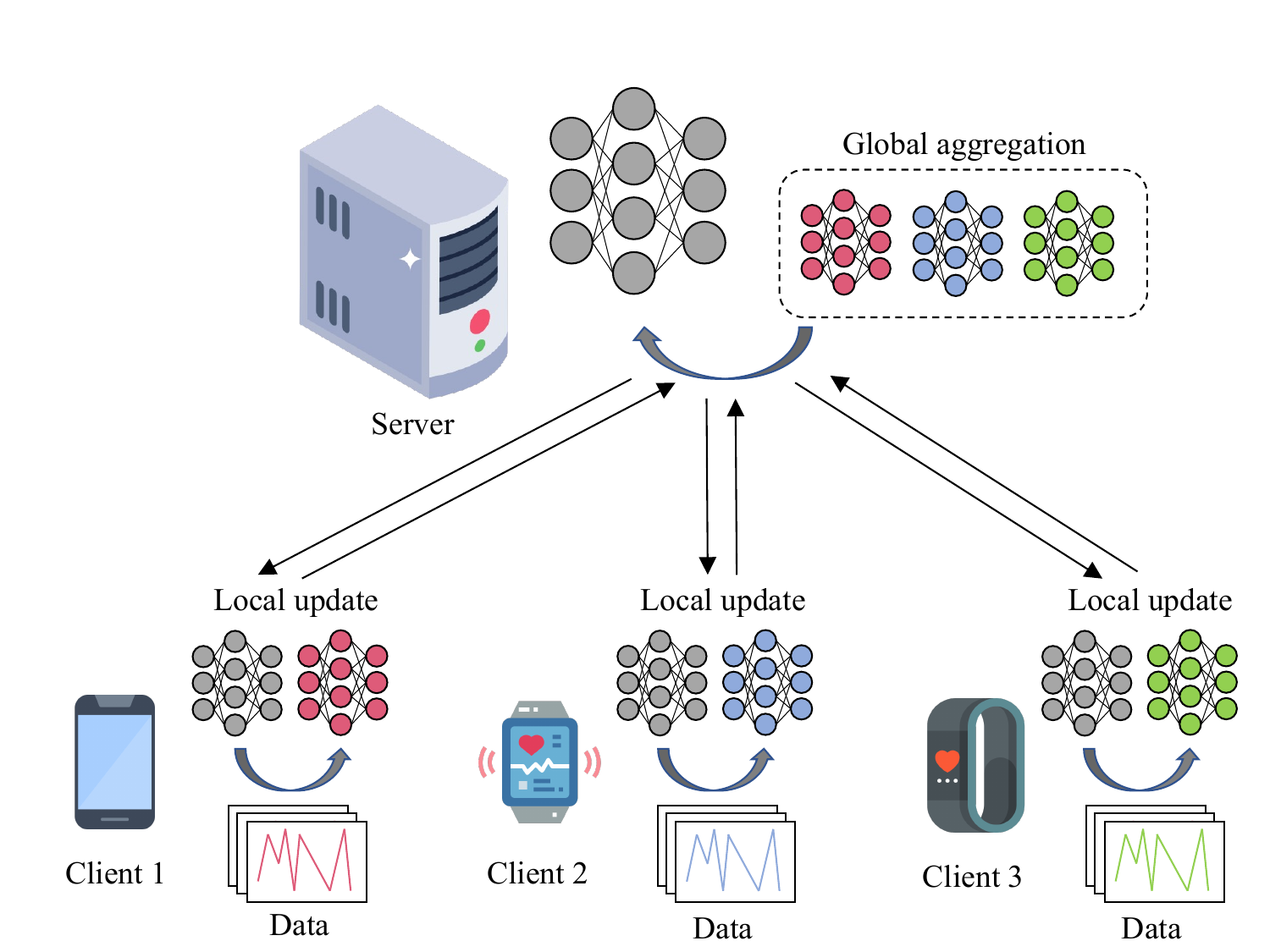}}
\caption{A overview of FL healthcare system across multiple edge devices, where participant clients receive model weights from the central server and execute local training, then send back the gradient updates, the server aggregates the updates and generates the model for the next round training.
\label{fig1}}
\end{figure}

The emergence of federated learning (FL), together with MEC and IoT, has provided a viable solution for smart wearable healthcare. In particular, FL allowed a collaborative learning scheme without data sharing \cite{b09}. Each party of FL trains the model using its own data and then sends the updated model parameters to a central server, where the weights are aggregated as the initial values for the next round \cite{b08}. FL protects user privacy fundamentally since the original data is always retained locally during the whole training process. Fig. \ref{fig1} demonstrates a typical structure of FL based healthcare system. MEC meanwhile, has provided ample computational power to the well-trained models running locally. As a result, detection results can be inferred and sent back to users within seconds by edge devices \cite{b15}.

For nowadays FL applications, one of the main challenges is that data from most clients usually own their own personalized styles and preferences, and this leads to heterogeneity among models trained by different clients \cite{b03}. The resulting diversity may not merely slow down the aggregated model convergence but also decrease the precision \cite{b11}. As a prime representative of data heterogeneity, the phenomenon of class imbalance is frequently seen in many real-world scenes, especially in healthcare and medicine fields \cite{b07}. Class imbalance refers to the situation when one class of a binary or multi-class classification problem has significantly fewer instances than the other class or classes, it is widely considered a fundamental and generic issue for health data. For example, even in patients with severe heart disease, their heart rate is abnormal for only a fraction of the whole duration, leading to majority class samples making up a significant amount of the total training data, whereas minority class samples account for a considerably smaller portion. The direct consequence of class imbalance is that models may ignore the accuracy of minority classes and concentrate on getting most samples processed correctly. However, in many practical scenarios, these minority classes are actually the targets that need to be found. Wearable healthcare devices also are expected to be more sensitive to abnormal situations than normal \cite{b15}\cite{b16}. Giving alarms in time for a sudden emergency is much more important than creating daily health reports.

In the structure of conventional machine learning, imbalanced distribution is not a big challenge. A number of approaches have been proposed to balance the ratio of majority and minority classes. However, in FL training processes, only the gradients and model parameters are allowed to be transferred between clients and the server. FL is always running with mechanisms to protect participant privacy and restrict the server from accessing any private information including distributions and classes ratio \cite{b17}. These mechanisms prevent users from addressing class imbalance with existing approaches of machine learning due to a lack of global knowledge. Data imbalance is hard to solve by clients alone since the mismatch between the local data distribution and the global distribution may impose an even more negative side-effect on the aggregated model \cite{b13}. Duan et al. proposed a solution that directly addresses the class imbalance among clients by collecting local data distributions of each client with proxy servers, which leads to a potential risk of exposing latent back doors to attack \cite{b20}. Zhang et al. pointed out that prior knowledge collected by the server also contains personal privacy while a client with more positive samples is more likely to correspond to a user with poor health, and then proposed a bottom-up method to schedule local updates by the client itself with deep reinforcement learning, which achieved further protection of privacy but the computational consumption of clients may become much higher \cite{b22}. Wang et al. designed a monitor scheme for class imbalance in FL that estimates selected data composition at each aggregation round, which addressed the imbalance issue without adding too much computation cost \cite{b23}. However, in practical scenarios of FL, prior knowledge or global distributions are not always set in stone. Hospitals tend not to wear a sensor on only one patient at all times, and a disease pandemic can increase the number of positive samples sharply and fall back in a short period of time. To our knowledge, there is no work that simultaneously protects participant privacy, including prior knowledge, and resolves distribution mismatch between the global server and client sides even when the mismatch keeps varying.

In our view, we considered FL for wearable healthcare as a task that is always in training. The users keep producing new data in their daily lives, so happening class imbalance is always a possibility. It is in some ways an online learning task that wearable devices may not have enough space to store too much historical data. Constantly updated and obsolete data makes it difficult to obtain stable prior knowledge of all the clients \cite{b19}. In this work, we tackled the mentioned challenges. We viewed each aggregation of the FL model as an opportunity to observe the global data distribution and designed an approach to estimate the sample composition of selected clients by comparing updated gradients of clients with gradients computed by auxiliary data in the central server. However, It is difficult to obtain accurate global knowledge for a single observation of selected clients, summarizing multiple observations in iterations and calculating the mean can improve the accuracy of the estimation but may induce the common model in the wrong direction in the event of a varying global distribution. In order to detect the dynamic class imbalance, we proposed an autoregressive ratio estimation function inspired by the Kalman filter to amplify the proportion of later observations in the average results while maintaining stability and accuracy \cite{b18} \cite{b21}.

In the proposed framework named FedImT (Federated Imbalance Track), inspired by the monitor scheme to class imbalance \cite{b23}, we first estimated the proportion of each class in the selected clients at each aggregation round of the FL model without asking clients to upload any private prior information. If the estimated ratio shows a significant mismatch with previous estimations, we would drop the gradients update this round but still conduct ratio observation iteration to track the changes in the global distribution. Finally, the model would be updated with a balanced Cross-Entropy loss with the observed composition knowledge, and the severe mismatch between the global and local data distribution is prevented from occurring during the entire update process. We compared FedImT with the advanced FL baselines for healthcare services. The results show that our approach achieves a significant performance against varying imbalanced data distribution while avoiding apparent extra computational overhead.

\section{Related Works}

\subsection{Mobile edge computing}

MEC is an emerging computing architecture that enables computing resources and digital services to be located closer to the client users at the edge of the whole network system \cite{b01}. As the computing performance of various edge devices increases, healthcare based on MEC and IoT technology has established itself as a component of everyday life \cite{b29} \cite{b24}. People can easily track their own health conditions with smartphones and other wearable devices. The body activities are collected from the sensors equipped on these smart devices and processed by the devices immediately. Participants should be informed right once if any abnormal symptoms are detected. MEC-based smart healthcare services have rapidly evolved in recent years. Latha et al. proposed a heart disease prediction model partially observable Markov decision process using fog computing \cite{b25}. Other examples include using a smartphone with an off-the-shelf WiFi router for human activity recognition with various scales \cite{b26}. In this work, we contribute to MEC-powered healthcare by designing an FL system for edge devices providing full privacy protection while training deep models for detecting abnormal symptoms and is robust to the class imbalance issue of health data even when the ratios of certain classes keep varying.

\subsection{Federated learning}

FL is an emerging distributed approach that allows multiple clients or participants to collaboratively train a model without sharing their private data with a central server \cite{b08}. Instead of sending data to a central location for storage and processing, the model is trained locally on each device, and only the updated model parameters are sent back to a central aggregator. FL allows data to be maintained on individual devices rather than being shared over a network, which makes it particularly helpful in circumstances where data privacy is an issue. Additionally, because each device can contribute its own distinct data to the training process, it enables the convergence of models on a wider and more varied set of data \cite{b03}.

Due to reliable privacy protection, FL is a nice fit for wearable healthcare where the training data consists of a large number of sensitive health conditions and personal biological signals of participants \cite{b06}\cite{b09}. A groundbreaking effort in this direction is FedHealth\cite{b30}, which developed an FL framework to aggregate models locally trained models from various sources and produced a personalized model using transfer learning. However, FedHealth was not designed for imbalanced environments where the ratio of positive samples may be extremely small. Another work that deserves to be highlighted is Astraea from Duan et al \cite{b20}. In Astrea, an approach to self-balancing data sampling that was committed to tackling the issue of class imbalance in order to increase model correctness and robustness. However, Astrea presupposed that the global server is aware of the class distributions of every participant, which exposes training data to various inference attacks and contradicted the participants' increasing requirements for privacy. FedSens is also a remarkable work where a bottom-up method was proposed to avoid uploading private information as well as prior knowledge to the central server \cite{b22}. In FedSens, the frequency of local updates is controlled by multi-agent deep reinforcement learning models on the client side. As a result, clients with more minority samples are instructed to execute more local updates as energy consumption allows. Although FedSens proved to be a significant progress compared to previous algorithms, its performance for scenarios with much more clients is not yet tested, and it is not able to respond instantly to variations in the global ratio. Consequently, it is essential to develop an imbalance-aware approach for identifying and addressing the class imbalance issue in FL but posing no threat to data privacy.

\subsection{Class imbalance}
In centralized machine learning, models often demand a large quantity of correctly labeled training data for optimizing their parameters. The problem of class imbalance is not a rare occurrence in many practical scenarios, e.g. object detection. Lin et al. analyzed the causes of one-stage detectors' accuracy inferior to two-stage detectors in the object detection field and pointed out it is due to class imbalance. We know that in the object detection process, an input image may contain thousands of candidate locations, but only a few of them include objects, which introduces class imbalance to model training. The authors then proposed Focal loss to reduce the gradient decent reward of major easy negative samples \cite{b32}. Prior methods to address class imbalance have basically followed one of the following philosophies: One is to increase the quantity of the samples as minority classes. such approaches leverage data resampling and data augmentation. DeVries et al. proposed a simple yet effective data augmentation technique called cutout, which involves randomly masking out rectangular sections of input images during training \cite{b27}. Deep reinforcement learning ways were also introduced to data augmentation by AutoAugment to search for the optimal augmentation policies for a given dataset and model architecture and achieved state-of-the-art performance on various image classification tasks \cite{b28}. Another idea is to relatively raise the cost of model misclassification for minority samples, which is also known as cost-sensitive learning methods. Cost-Sensitive Deep Feature Representation is one of the representative cost-sensitive learning works in recent years which proposed a two-stage approach to extract cost-sensitive representations for balanced training \cite{b34}. The first is a data-level solution to class imbalance while the other addresses this issue by modifying algorithms and models. Of course, there are also approaches that solve imbalances from both data and algorithm aspects.

However, due to the privacy concern of FL, few of the existing techniques can perform well in FL scenes. Obviously, we are prohibited from oversampling or copying data across the clients and those totally healthy participants may have absolutely no abnormal conditions recorded on their devices. Loss functions like Balanced Cross Entropy may help mitigate the class imbalance problem but it requires knowledge of the overall ratio \cite{b40}, which may threaten privacy. Allowing each client to train separately in a balanced manner is likely to lead to a severe mismatch. In our opinion, in order to hedge against the risk of privacy leakage, participants are forbidden to share either raw data or ratio information with other parties (including the server, the proxy server, and other participants). This strict requirement for privacy demands new approaches to address imbalance issues.

\section{Methodology}
In this section, we first formulate the problem of FL based wearable healthcare system, our expectation is that it can perform well towards varying imbalanced distributed data while meeting the privacy requirements of participants. The system is defined by a standard FedAvg structure and procedure \cite{b08}. After that, we presented our method to estimate the composition ratio of data from selected clients in a single aggregation round, and then introduced an autoregressive function to track its variations. We update this global model using a dynamic re-weighted balanced Cross-Entropy loss. The system overview is presented in the last subsection.

\subsection{FedAvg}
FedAvg has been known as the de facto algorithm of FL and most of the subsequent proposed FL methods are actually its variants \cite{b42}. In each round of FedAvg updates, the algorithm can be divided into three steps. First, the server shares the global model with the clients as initial weights. Then, the clients train their own models for certain epochs with local data and send back their updates. At last, the server aggregates the received weights based on the sample quantity of each client, to generate an updated global model for the next round \cite{b08}.

In essence, FedAvg aims to solve a minimization of a global objective function $L_{g} \left ( \omega  \right ) $:

\begin{equation}
\min L_g\left ( \omega  \right ) =\sum_{k=1}^{K} p_kL_k \left (w  \right ),
\label{eq1}
\end{equation}

\noindent where \emph{K} means the number of participant clients in the FL system, \emph{k} stands for the index of each client, \emph{$p_k$} denotes the ratio of the \emph{k}-th client's samples to the total samples of all clients so that it must satisfy $\sum_{k=1}^{K}p_k=1 $. $L_k\left ( \omega  \right )$ stands for the local loss of each client, which is defined as:

\begin{equation}
L_k\left (\omega\right) = \mathbb{E}_{\left(x,y\right)\sim\upsilon\left (x,y\right)}L\left(f_k\left(x,\omega\right),y\right).
\label{eq2}
\end{equation}

\noindent In equation \ref{eq2}, $\upsilon\left (x,y\right )$ stands for the joint distribution of data in client \emph{k}, and \emph{f} stands for the classifier model in client \emph{k}. \emph{L} is the loss function. Besides the basic iteration, FedAvg also offers some hyperparameters to control the aggregation strategy. Considering the efficiency issues, most systems of FedAvg do not aggregate the weights of all clients at each round, but randomly select clients at a defined rate $\eta$:

\begin{equation}
\frac{\left | S_j \right | }{\left | S \right | } =\eta,
\label{eq3}
\end{equation}

\noindent with $S_j$ being the set of select clients at epoch $j$ and $S$ being all clients.

In addition, to reduce communication costs and improve the robustness of aggregated gradients, \emph{E} is usually set to represent the number of local training epochs between two aggregations. it can also be considered as an aggregation interval. This local learning process between aggregations for clients is referred by us to as a local update:
\begin{equation}
\omega _{k}^{j+1} \gets \omega _{k}^{j} + \alpha \bigtriangledown L_k\left (\omega _{k}^{j}  \right ) ,j \bmod E \neq 0.
\label{eq4}
\end{equation}

\noindent The aggregation process of the server is referred to as a global aggregation:

\begin{equation}
\omega ^{j+1} =\sum_{k\in S_j }^{} p_k\omega _{k}^{j}, j \bmod E=0.
\label{eq5}
\end{equation}

\noindent in which $\omega_{k}^{j}$ stands for \emph{k}-th update of client \emph{k} at epoch \emph{j}, and $\alpha$ represents the learning rate. In general, the learning rate is consistent across clients.

\begin{figure}[!h]
\centering{\includegraphics[width=\linewidth]{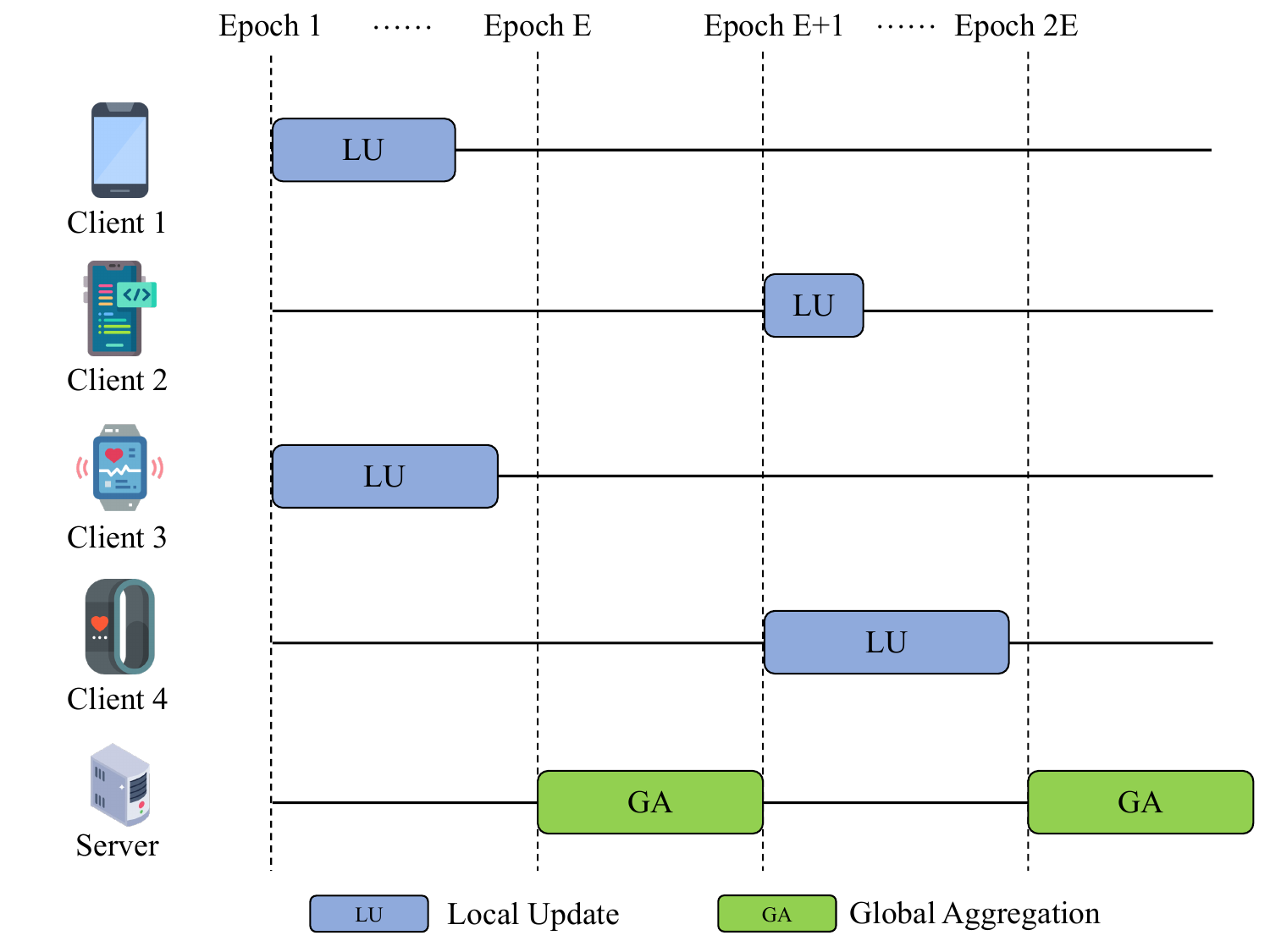}}
\caption{Workflow of FedAvg algorithm in a healthcare system where randomly selected edge devices asynchronously perform local updates while the server aggregates updates in a defined frequency.
\label{fig2}}
\end{figure}

We illustrated the workflow of FedAvg in an FL healthcare system in Fig. \ref{fig2}. The system consists of four slave workers and a master server. These workers including common personal digital devices or wearables, play the role of clients in FL. We assumed that these slave workers are equipped with various necessary sensors and capable of performing local training on their own collected data. The master server can be deployed with any nearby node in the network such as base stations, smart routers, or even gateways. The server in FL does not have a strong need for computing ability for its low computational overhead, and more often than not, lower latency and more reliable connections matter more for the server.

\subsection{Composition ratio estimation}

In order to train the model in a balanced manner, it is necessary to estimate the composition ratio of data from selected clients at each aggregation round. we formulated mentioned class imbalance issue on a multi-class forward neural network $f$ whose last layer is a fully connected dense layer with a softmax operation so that we could make sure the output size of the deep classifier equals the number of classes $Q$. Here the label space of the training task could be defined as $Y=\left\{ 1,2,..., Q \right\} $. To cover more scenarios with different structures of models, we represented all hidden layer computations as a standalone function $H$. In the training procedure, the $i$-th sample of class $p$ was denoted as $X_i^p$. When $X_i^p$ was fed into the neural network, the corresponding output of the deep classifier without softmax operation was denoted as $O_i^p=\left[o_{i, (1)}^p,\cdots, o_{i, (Q)}^p\right]$, the output of $H$ was denoted as $H(X_i^p)=\left[h_{i, (1)}^p,\cdots, h_{i, (s)}^p\right]$ where $s$ was the neuron number of the last hidden layer. We used $\omega$ to define the overall parameter of the whole deep model. In particular, $\omega(j)$ was the parameter in the $j$ training epoch. Moreover, the linear connected weights between $H$ and the output layer were denoted as $W=\left[W_{(1)},\cdots,W_{(Q)}\right]$, and $W\in\omega$. According to the calculation principle of dense layers, we found that for $X_i^p$, its corresponding output for class $q$ satisfied $W_{(q)}\cdot  H\left(X_i^p\right) = o_{i,(q)}^p$, and $q\in\left\{1,\cdots, Q\right\}$. In each training epoch, back-propagation was used to calculate the gradient of prediction loss $L\left(\omega\right)$ subject to $\omega$. We introduced $\alpha$ to represent the learning rate, then the local update procedure could be written as:

\begin{equation}
\omega(j+1)=\omega(j)-\alpha\frac{\partial L(\omega(j))}{\partial \omega(j)}.
\label{eq6}
\end{equation}

In the present deep learning methods, samples are not fed into the model one by one, but using mini-batch training. All data is divided into small batches of a fixed and defined size, each batch is applied to update the model parameters in the direction of minimizing prediction errors. In the vast majority of cases nowadays, the batch size is typically chosen to be a power of 2, such as 32, 64, or 128, to take advantage of the parallel processing capabilities of GPUs \cite{b41}. As GPU productivity has continued to improve over the years and memory has increased dramatically, there has been a trend towards larger batch sizes.

In the backpropagation of error gradients in mini-batch training, the first component to be updated is the connection weights $W$ between the output layer and the last hidden layer. The gradients can be written as:

\begin{equation}
\triangle W_{batch}=-\frac{\alpha}{bs}\sum_{p=1}^{Q}\sum_{i=1}^{n_p}\frac{\partial L(\omega)}{\partial W_{(p)}},
\label{eq7}
\end{equation}

\noindent where $n_p$ denotes the number of samples for class $p$ in this mini-batch, and $bs$ is the batch size.

In some of the frontier empirical studies of deep learning, researchers observed that the samples of the same class $p$ were expected to result in similar outputs of hidden layers $\left[h_{i, (1)}^p,\cdots, h_{i, (s)}^p\right]$, the larger batch could even help further reduction of incidental cases for similar outputs \cite{b23}. And inside each mini-batch, there is no update operation, thus the corresponding gradients could be also close to a fixed value, we assumed that the fixed value is the mean of gradients calculated by samples of class $p$, and denoted it as $\nabla L\left(\omega, W_{(p)}\right)$, then (\ref{eq7}) could be written as:

\begin{equation}
\triangle W_{batch}=-\frac{\alpha}{bs}\sum_{p=1}^{Q}\left(\nabla L\left(\omega, W_{(p)}\right ) n_p\right ).
\label{eq8}
\end{equation}

The next step was to expand $\triangle W$ to epoch level, here we assumed that within each training epoch, the parameter change is relatively small and could be ignored. Under this circumstance, $\nabla L\left(\omega, W_{(p)}\right)$ of different mini-batches within an epoch remains the same. Therefore, we could work out $\triangle W_{epoch}$ by accumulating $n_p$ in all mini-batches, as:

\begin{equation}
\triangle W_{epoch}=-\frac{\alpha}{bs}\sum_{p=1}^{Q}\left(\nabla L\left(\omega, W_{(p)}\right) N_p\right),
\label{eq9}
\end{equation}

\noindent where $N_p$ denotes the total quantity of class $p$.

In the most common setting of FL (FedAvg), the local updates of selected clients are aggregated with corresponding weights of data amount. That means, as with the expansion from batch-level to epoch-level, we could get the global gradient update for $W$ by accumulating $N_p$ of all selected clients. it was written as:

\begin{equation}
\begin{aligned}
\triangle W_{global}^{(p)}&=-\frac{\alpha}{bs}\eta\sum_{k=1}^{K}\left(\nabla L\left(\omega, W_{(p)}\right) N_p\right)\\
&= -\frac{\alpha}{bs}\eta\nabla L\left(\omega, W_{(p)}\right)\left ( \sum_{k=1}^{K} N_p\right ),
\end{aligned}
\label{eq10}
\end{equation}

\noindent where $K$ was the number of selected clients, and $K=\frac{1}{\eta}$, $\triangle W_{global}^{(p)}$ was the gradients induced by all samples of class $p$ from selected clients during this aggregation, $\left(\sum_{k=1}^{K} N_p \right) $ was actually the target value we needed to estimate while aggregations to achieve balanced training, which meant overall sample quantity of class $p$ in selected clients.

Based on the above analysis, we designed our method to estimate the composition of data as follows. During each time aggregation of round $j+1$, the server will individually input the auxiliary data of each class to the global model of the previous iteration, denoted as $G_t$. Then it produces the corresponding gradient update for connection weights $\left\{\triangle W_{aux}^{(1)},\cdots,\triangle W_{aux}^{(Q)}\right\}$, where each $W_{aux}^{(p)}$ corresponds to class $p$. For the corresponding output node for class $p$, its connection weight value with the $m$-th node in the last hidden layer was denoted as $W_{(p,m)}$, and the corresponding gradient update induced by auxiliary data of class $q$ was defined as $\triangle W_{(p,m)}^{(q)}$. According to the principle of gradient backpropagation, we formulated the accumulation of weight changes in the setting of standard FedAvg. That was, for the connection weight between the output node of class $p$ and the $m$-th node in the last hidden layer, its weight updates satisfied:

\begin{equation}
\begin{aligned}
&\triangle W_{(p,m)}^{(p)}\cdot\sum_{k=1}^{K}\left(N_p\right)+\triangle W_{(p,m)}^{(\neg p)}\cdot\sum_{k=1}^{K}\left(N_{\neg p}\right) \\
&=N_{aux}^{(p)}\cdot K \cdot \left(W_{(p,m)}^{j+1}-W_{(p,m)}^j\right),
\end{aligned}
\label{eq11}
\end{equation}

\noindent where $N_{aux}^{(p)}$ denoted the amount of auxiliary data for class $p$, $\sum_{k=1}^{K}\left(N_p\right)$ was the target of our estimation, $W_{(p,m)}^{j+1}$ and $W_{(p,m)}^j$ were the connection weights of $W_{(p,m)}$ in round $j+1$ and round $j$, $\triangle W_{(p,m)}^{(\neg p)}$ meant the expected gradient update of non $p$ classes for $W_{(p,m)}$, and it was calculated as:

\begin{equation}
\triangle W_{(p,m)}^{(\neg p)}=\frac{\sum_{q=1}^{Q} \triangle W_{(p,m)}^{(q)}-\triangle W_{(p,m)}^{(p)}}{Q-1},
\label{eq12}
\end{equation}

\noindent and $\sum_{k=1}^{K}\left(N_{\neg p}\right)$ denoted the overall sample amount of non $p$ classes, which could be written as:

\begin{equation}
\sum_{k=1}^{K}\left(N_{\neg p}\right)=\sum_{k=1}^{K}\sum_{q=1}^{Q}N_q-\sum_{k=1}^{K}N_p,
\label{eq13}
\end{equation}

\noindent in which $\sum_{q=1}^{Q}N_q$ was the overall sample quantity on client $k$. $\sum_{q=1}^{Q}N_q$ was the only parameter that clients needed to upload, which would also have been collected by the server to calculate the weight of each client in aggregations, demonstrating little privacy risk.

Then, all values except $\sum_{k=1}^{K}\left(N_p\right)$ could be acquired in the above equations. We could calculate $\sum_{k=1}^{K}\left(N_p\right)$ with gradient updates of connection weights between the output node for class $p$ and any one of the nodes in the last hidden layer. For a clearer presentation, we denoted $\sum_{k=1}^{K}\left(N_p\right)$ value calculated by updates with $m$-th hidden node as $\widehat{N}_{p,m}$. However, in practice, not every linear connection towards the node of class $p$ appeared to have a significant weight update. We considered that the weights whose updating magnitudes are relatively large were likely to lead to an accurate estimation of sample quantity so we designed the aggregation of $\widehat{N}_{p,m}$ as a weighted summation, the component weight $C_{(p,m)}$ of each $\widehat{N}_{p,m}$ was calculated as:

\begin{equation}
C_{(p,m)}=\frac{\left(Q-1\right)\triangle W_{(p,m)}^{(p)}}{\sum_{q=1}^{Q}\triangle W_{(p,m)}^{(q)}-\triangle W_{(p,m)}^{(p)}}.
\label{eq14}
\end{equation}

\noindent And the final $\widehat{N}_p$ could be written as:

\begin{equation}
\widehat{N}_p=\sum_{m=1}^{s} \frac{C_{(p,m)}}{\sum_{m=1}^{s} C_{(p,m)}} \widehat{N}_{(p,m)}.
\label{eq15}
\end{equation}

After the estimations for all classes were completed, we could work out the vector $\left[\widehat{N}_1^{j},\cdots \widehat{N}_Q^{j}\right]$ describing the data composition at round $j+1$, and only needed clients to upload their overall sample amount, which was also the basic requirements of FedAvg and not privacy sensitive \cite{b08}. In addition, we also offered another option to avoid this only requirement by setting all the clients to train on their latest $N_{latest}$ data, $N_{latest}$ was the number of latest data obtained for training in each client. However, abnormal states of physical health conditions usually appear in periods, which means the distribution of samples is not uniform in the time dimension. Only training with a fixed size of the latest data may lead to the composition ratio for all classes of training data keeping varying strongly. The following content of this paper proposed an autoregressive function to help track the varying imbalance of data and solved the issue by re-balancing the weights of the loss function before each FL training round.

\subsection{Autoregressive ratio observation}
After illustrating the method to execute composition ratio estimations, we proposed a concept that we could avoid sharing any local information with the server by limiting the training data size to the latest $N_{latest}$ samples. There are some obvious benefits to this setting. recently collected data are more relevant and representative of the current state of participants, which meets our demand for more responsive detections. Moreover, we considered an FL healthcare system as a scheme that is always in training, where new data is constantly collected by clients. In this way, training with the entire data actually lets old data join more local updates in the whole training procedure, which may hurt robustness and precision.

However, the non-iid (not identically and independently distributed) distribution of data may lead to various challenges in such an FL setting. On the client side \cite{b11}, the happening distribution of abnormal health conditions is not uniform in the timeline. Once a person is detected with an accelerated heart rate, the painful and nervous condition usually lasts a while before ending, and so are human activities and emotions. This distribution characteristic can make the class ratio of the latest samples keep varying in different FL rounds. On the server side, due to the individual heterogeneity among participants, the composition ratio of selected clients' data may contain a significant mismatch with the global. Therefore, the inferred ratio knowledge of round $j$ cannot be referenced in round $j+1$. Based on the above analysis, addressing the varying imbalance issue is essential to well-train an FL healthcare system in the proposed privacy protection setting.

Since we already had the capability to estimate the data composition of each aggregation, we decided to solve the issue by summarizing more estimates. At the beginning, we converted the obtained quantity vector $\left[\widehat{N}_1^{j},\cdots \widehat{N}_Q^{j}\right]$ to ratio form as:

\begin{equation}
R_p^j=\frac{N_p}{\sum_{q=1}^{Q}Nq},
\label{eq16}
\end{equation}

\noindent where $R_p^j$ denoted the ratio of class $p$ in training data composition at round $j$.

Due to that most FL servers will not collect all the clients' updates in every aggregation. There may be heterogeneity between the selected and the global. From general experience, as the training process progresses, taking multiple measurements and calculating an average value can help improve accuracy. However, the ratio inside each client also faces variations. Average calculation obviously does not take into account the underlying dynamics. Here we introduced the autoregressive ratio observation function.

Firstly, we rewrote the average calculation into a form of recursion function:

\begin{equation}
\begin{aligned}
\widehat{R_p^j} &= \frac{1}{j}\cdot\left(R_p^1+R_p^2+\cdots+R_p^j\right)\\
&=\frac{1}{j}\cdot\left(R_p^1+R_p^2+\cdots+R_p^{j-1}\right)+\frac{1}{j}\cdot R_p^j\\
&=\frac{1}{j}\cdot\frac{j-1}{j-1}\cdot\left(R_p^1+R_p^2+\cdots+R_p^{j-1}\right)+\frac{1}{j}\cdot R_p^j\\
&=\frac{j-1}{j}\cdot\frac{\left(R_p^1+R_p^2+\cdots+R_p^{j-1}\right)}{j-1}+\frac{1}{j}\cdot R_p^j\\
&=\frac{j-1}{j}\cdot\widehat{R_p^{j-1}}+\frac{1}{j}\cdot R_p^j,
\end{aligned}
\label{eq17}
\end{equation}

\noindent where $\widehat{R_p^j}$ stood for the inferred ratio value for class $p$ at round $j$ by average calculation.

According to (\ref{eq17}), we found that with $j$ increases, the new estimated ratio $R_p^j$ contributes little to the value of $\widehat{R_p^j}$, which causes $\widehat{R_p^j}$ fails to track the system dynamics or the characteristics of the observation noise. However, only considering the value of $R_p^j$ is either not reliable for partial observation. In order to accurately respond to variations in data composition while keeping historical knowledge, we modified the coefficients in front of the recurrence term and the new coming observations. It could be written as:

\begin{equation}
\widehat{R_p^j}=\left\{
\begin{aligned}
    &R_p^1,\quad j = 1 \\
    &\frac{1-\eta}{2} \cdot \widehat{R_p^{j-1}} +\frac{\eta}{2}\cdot R_p^j,\quad j > 1
\end{aligned}
\right.
\label{eq18}
\end{equation}

\noindent in which $\eta$ denoted the ratio of selected clients within all participant clients as aforementioned, it was believed by us to be quite a suitable gain factor of this autoregressive function. Moreover, setting the gain factor with $\eta$ can be effectively integrated with scenarios of the dynamic ratio of selected clients. This approach follows the logic that the more clients it selects, the more important the estimation counts.

Based on the above method, we could track the varying imbalanced ratio of data with emphasis on later observations and reference to historical knowledge. Accordingly, we had a chance to re-balance the weighted Cross-Entropy loss function for the next round after aggregation. The balanced Cross-Entropy loss in this work was defined as follows:

\begin{equation}
CB^{j+1}(y,\widehat{y_p})=\frac{1-\beta}{1-\beta^{N_{latest}\cdot \widehat{R_p^j}}}\cdot CE(y,\widehat{y_p})
\label{eq19}
\end{equation}

\noindent where $CB^{j+1}$ denoted the class-balanced loss function we planned to broadcast to clients for local updates in round $j+1$, $y$ and $\widehat{y_p}$ were predicted probabilities and ground truth value for class $p$, $\beta$ was applied as a hyperparameter, and $y\in[0,1)$, CE stood for common Cross-Entropy function for multi-class classification. Obviously, the $N_{latest}$ scheme is just an option. If collecting $\sum_{k=1}^{K}\left(N_p\right)$ is possible, we can get even higher performance by replacing $N_{latest}\cdot \widehat{R_p^j}$ with $\sum_{k=1}^{K}\left(N_p\right) \cdot \widehat{R_p^j}$ in (\ref{eq19}). It depends on the privacy requirements of users.

\subsection{Summary of FedImT workflow}

\begin{figure}[!t]
\centering{\includegraphics[width=\linewidth]{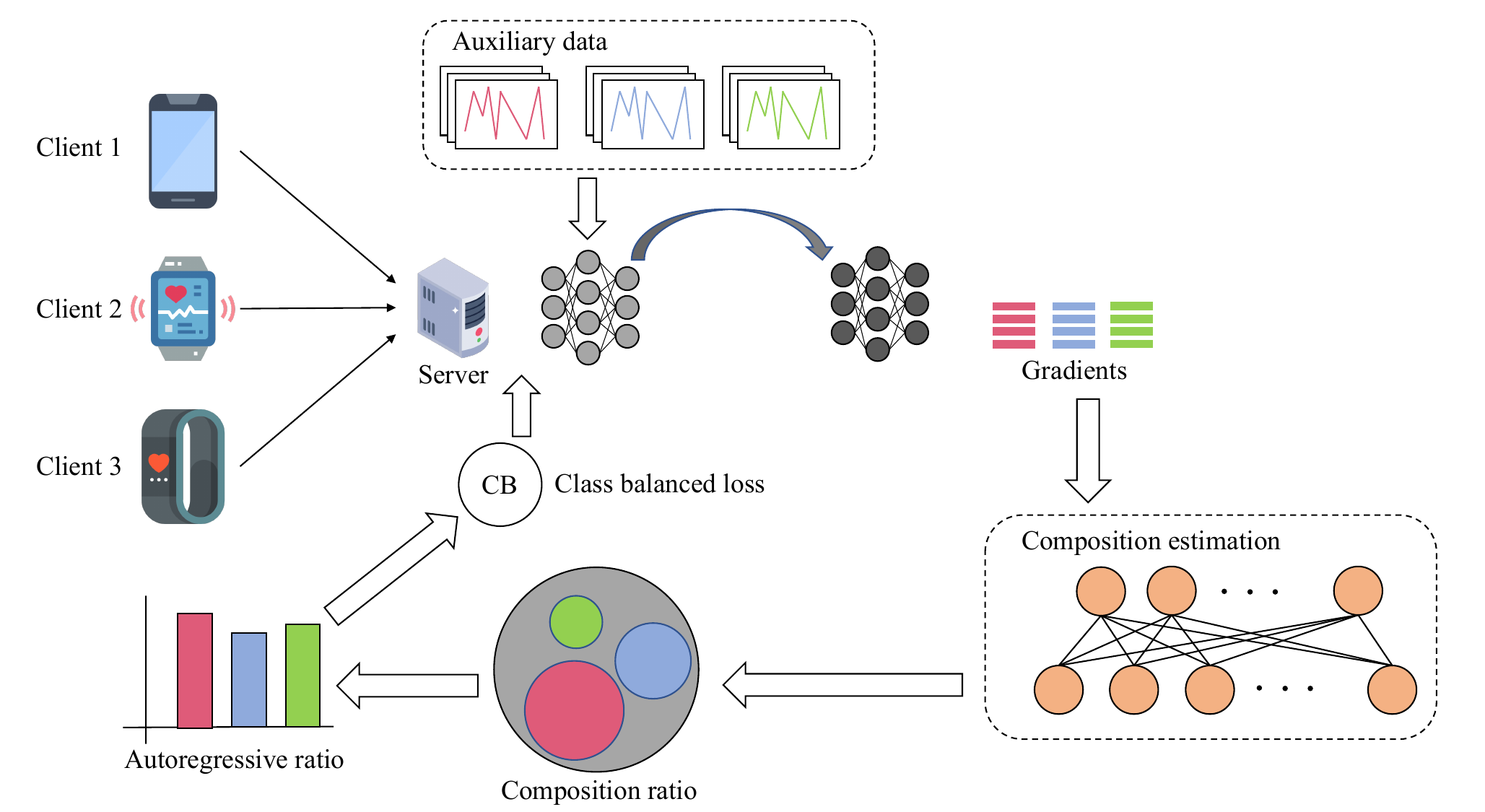}}
\caption{A overview of FedImT framework, the server infers the composition ratio of training data based on updates of connection weights with the assistance of auxiliary data, the ratio knowledge will be updated in a self-regressive way, and the loss function will be re-balanced after aggregation at each round.
\label{fig3}}
\end{figure}

Finally, we summarized the FedImT framework in Fig. \ref{fig3}. In the preliminary phase, the server initializes the ratio for each class with $\frac{1}{Q}$, and the class-balanced loss function also stays balanced among classes. At the round $j+1$, the server first collects the trained models from selected clients and executes aggregation, then auxiliary data will be fed into the aggregated model class by class. An estimation of training data composition will be made according to the gradients update of the dense connections in the model and merged with historical ratio information in a recursive way. After that, the loss function of all participant clients will be balanced with autoregressive ratio knowledge from round $1$ to round $j$ and get ready for round $j+1$ local training.

\section{Experiments}

In this section, we presented an extensive empirical evaluation of the FedImT approach using both standard machine learning benchmark testing data and data traces collected from a real-world healthcare application. The experiment result shows that FedImT achieves significant and stable performance advances in terms of the FL model training while reducing energy consumption compared to other leading FL methods.

\subsection{Case instruction and datasets}

Three datasets were introduced to our experiment evaluation: \textbf{MNIST}\cite{b31}, \textbf{UCI-HAR}\cite{b39}, and \textbf{Stay Alert! The Ford Challenge}\cite{b36}. MNIST is the most widely used benchmark dataset in machine learning, especially computer vision scenarios. UCI-HAR is a sensor dataset collected by smartphones put on the volunteer's waist, recording six activities like WALKING, WALKING(UPSTAIRS), WALKING(DOWNSTAIRS), SITTING, STANDING and LAYING. The Ford Challenge represents a rather practical scenario where wearable sensors are applied to detect driver drowsiness. The Ford Challenge contains a training set of $604,329$ samples and a test set of $120,840$ instances. In this application, we noted a significant class imbalance where only $16.2\%$ of the training dataset is of positive class labels.

\subsection{Baselines}
To our knowledge, we find that there are seldom existing methods designed to address the varying imbalanced distribution of non-iid data while satisfying strict privacy requirements and respecting the computation resource of edge devices. Therefore, we choose the following comparisons that are representatives of FL methods which are more likely to be applied to healthcare applications. They are: \textbf{Standard FedAvg}\cite{b08}, \textbf{FedProx}\cite{b37}, \textbf{FedNova}\cite{b38}, \textbf{FedHealth}\cite{b30}. Among these baselines, standard FedAvg is the prototype of almost all the existing FL approaches. FedProx is an extension of the standard FedAvg which introduces an additional regularization term to the training objective. FedNova is a normalized averaging approach that preserves quick error convergence while eliminating objective inconsistency was created and proposed. FedHealth is considered one of the state-of-the-art FL approaches that deploy the traditional FedAvg algorithm for distributed training of health data. It introduces transfer learning operations to traditional FL frameworks to build personalized models and enable accurate and personalized healthcare service without compromising user privacy and information security.

\subsection{Experiment setup}

We implemented our FedImT algorithm as well as other comparisons with the PyTorch library (version 1.13). our experiment built a federated learning process following the standard structure of FedAvg. We chose three different datasets for evaluation: MNIST, UCI-HAR, and The Ford Challenge, each of the three datasets has its own tendency for testing.

First of all, MNIST has been widely considered a general task for benchmarking and evaluating machine learning algorithms, particularly in the context of image classification tasks. MNIST is always being used as a starting point for testing and comparing different models due to its simplicity and quick experimentation feature. In order to introduce class imbalance and data heterogeneity to MNIST, we first grouped all the samples by their labels, then divided them into shards, and subsequently, these shards were randomly dispatched to a variety of clients. We noticed that this operation caused a significant class mismatch among clients. Due to that FL systems often chose only a part of participant clients for aggregation each round, and the data composition of each round was also of obvious imbalance.

For each case study, we implemented and utilized the following deep learning networks: LeNet5 for MNIST, a 6-layer convolutional neural network for UCI-HAR, and a 3-hidden-layer multilayer perceptron model \cite{b33} for The Ford Challenge. The local training batch size was 32, and the learning rate $\alpha = 0.001$, the gradient optimizer was Stochastic Gradient Descent (SGD), with momentum acceleration. We would train the MNIST task for 50 rounds, with 10 local rounds. As for the UCI-HAR and The Ford Challenge, the global round would be up to 80 rounds for better convergence. The auxiliary data was obtained by randomly selecting a number of samples of each class with putting-back from the training set. In practical scenarios, it can be acquired from public data or synthesized by generative models. If the training data for the FL system as a whole does not change considerably, this auxiliary data can be used for a long period. The size of the auxiliary data set in our experiments was 4 batches. In the MNIST task, we created 50 clients for FL training while the number of participants would be consistent with raw data in the other two case studies.

In order to clearly focus the testing target on the class imbalance issue, we rebuilt the standard setting of the UCI-HAR dataset where every participant would contribute to both the training set and the testing set, instead of training with a part of the volunteers and evaluating with the others. Considering the UCI-HAR dataset itself does not have a significant class imbalance, we enabled the $N_{lates
t}$ scheme where a sliding window was applied and only half of the data on each client would be chosen for local updates as the training process went on. When the $N_{latest}$ scheme is enabled, the learning rate will be set to $\alpha = 0.002$.

\subsection{Estimation performance}

\begin{figure}[!h]
\centering{\includegraphics[width=0.8\linewidth]{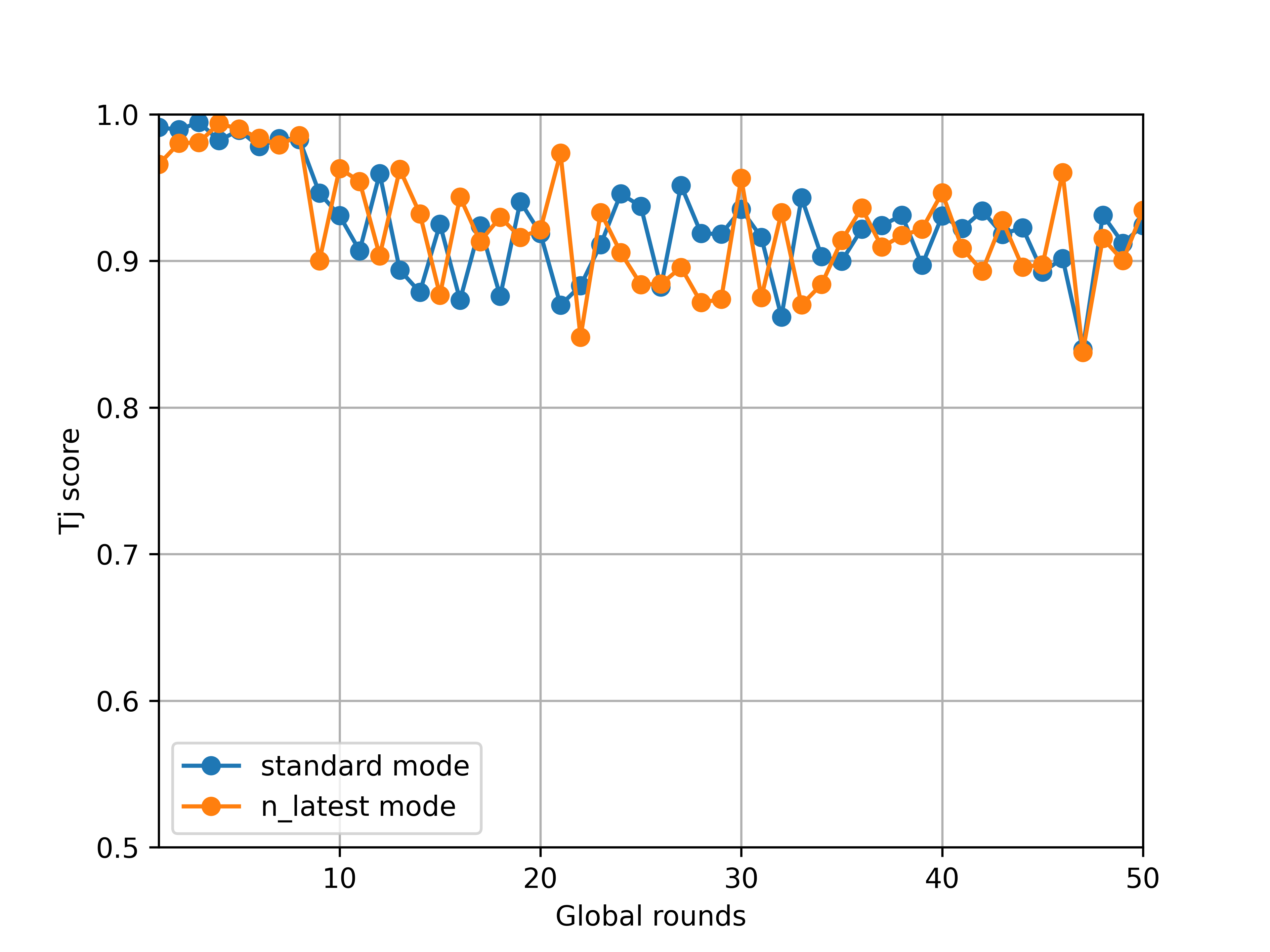}}
\caption{Cosine similarity metric between ratio estimation of data composition and ground truth each round.
\label{fig4}}
\end{figure}

\begin{figure}[!h]
\centering{\includegraphics[width=0.8\linewidth]{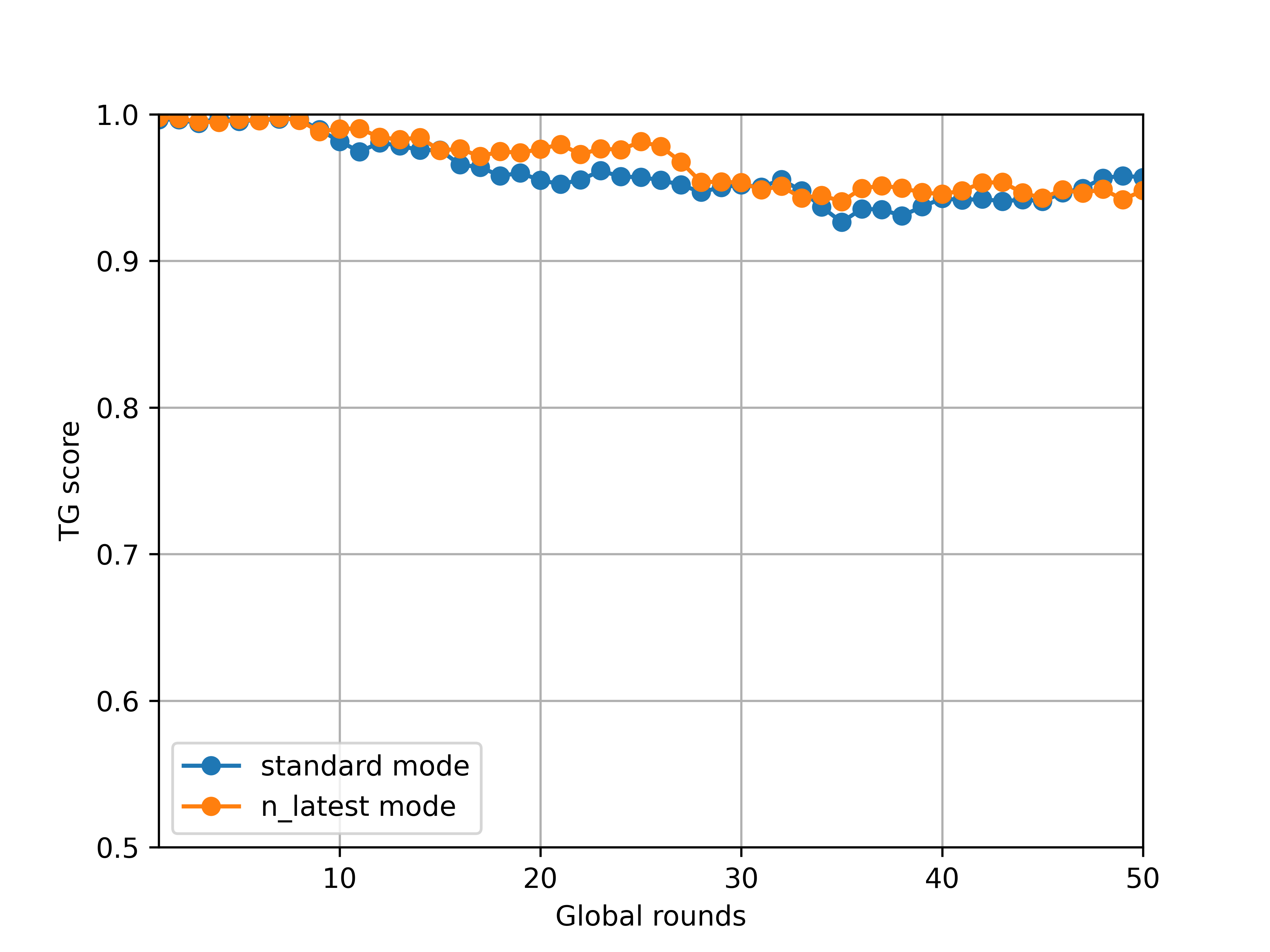}}
\caption{Cosine similarity metric between our autoregressive ratio observation of global data composition and ground truth in the whole training process.
\label{fig5}}
\end{figure}

So as to evaluate the performance of our model to estimate the data composition each round and over the whole training process, we planned an experiment where the aggregation server would randomly select 15 clients ($\eta = 0.3$) from 50 participants each round in MNIST task training. Every round contained 5 local epochs and there were 50 global rounds in total. For each participant client, the number of classes they owned was determined from 1 to 10 randomly. 3 is the number of plurality. As already explained in detail in the earlier section, the model was expected to compute an estimation vector $\left[R^j_1, R^j_2, ..., R^j_p\right]$ in the global round $j$. We compared it against the ground truth data composition with a measure of the cosine similarity score. This similarity metric was defined as $T_j$. The closer $T_j$ was to $1.0$, the more accurate our estimation in each round was. Moreover, we set another cosine similarity score that the autoregressive ratio estimation was compared against global data composition, defined as $T_G$. We recorded the evolution of $T_j$ and $T_G$ in both standard FL mode and $N_latest$ mode. Fig. \ref{fig4} and Fig. \ref{fig5} demonstrated the trends of $T_j$ and $T_G$ in the whole training process.

As is shown in Fig. \ref{fig4}, we can observe that our ratio estimation of training data was actually very close to the ground truth in every single round. In our case study, among both standard mode and $N_{latest}$ mode, the average $Tj$ score was above 0.9 and kept higher than 0.92 most of the time. Such results showed the effectiveness and precision of our estimation method. Moreover, introducing autoregression to our ratio observation can further improve the stability of convergence. Fig. \ref{fig5} demonstrated that autoregressive observation can significantly raise the expectation, especially the floor bound of estimation performance. All the values in our experimental training reached a higher score than 0.92, while the average value was over 0.95 in the whole process. After that, we also carried out some experiments with other hyperparameters and found that the expectation of $T_j$ and $T_G$ scores are positively correlated with the number of selected clients during aggregation. We also found that larger batches and more local epochs may cause a slightly positive impact on the estimation performance of our model.

\begin{figure*}[!t]
\centering{\includegraphics[width=0.8\linewidth]{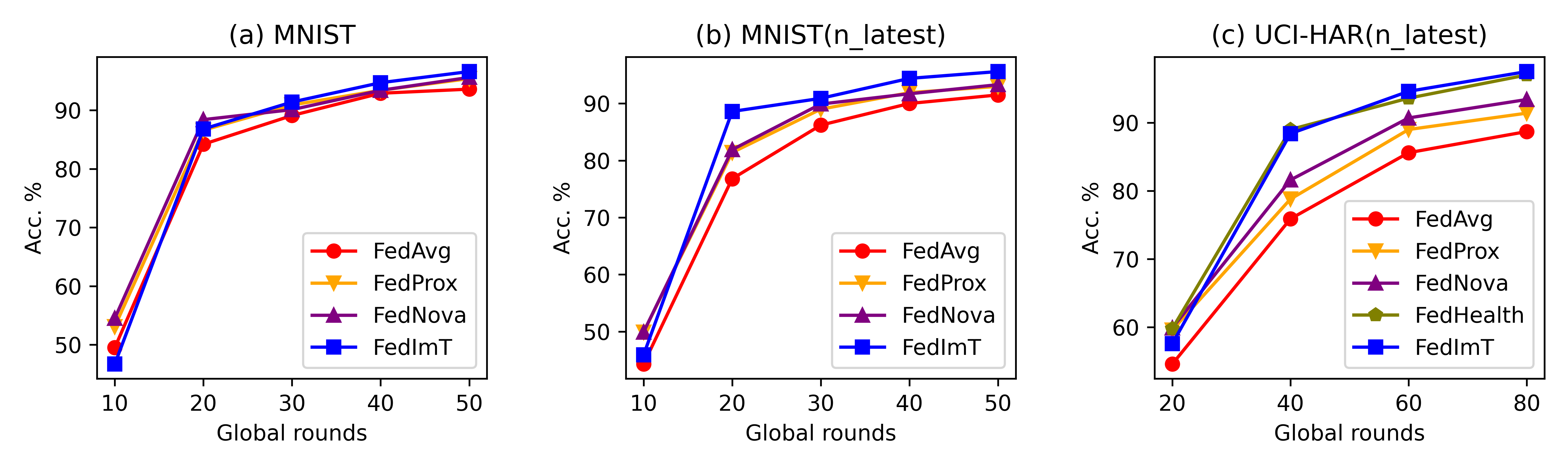}}
\caption{Comparison between our method (FedImT) and previous methods (FedAvg, FedProx, FedNova, and FedHealth), over the MNIST and UCI-HAR datasets, with different stages of the training process.
\label{fig7}}
\end{figure*}

\begin{figure*}[!t]
\centering{\includegraphics[width=0.8\linewidth]{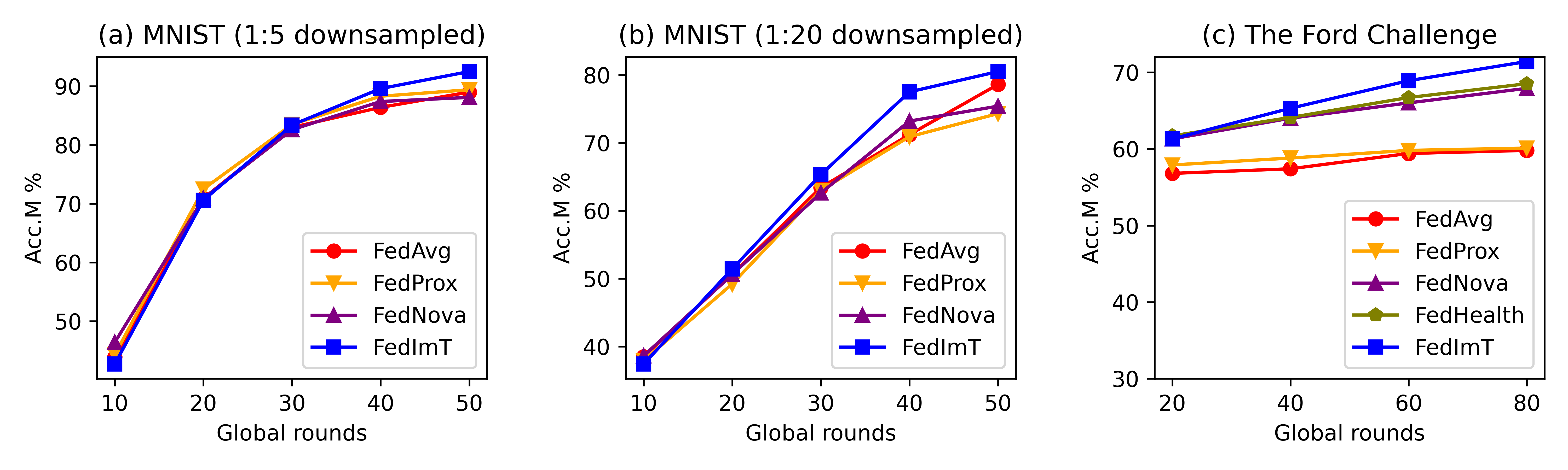}}
\caption{Comparison between our method (FedImT) and previous methods (FedAvg, FedProx, FedNova, and FedHealth), over the downsampled MNIST and the Ford Challenge case studies, with different stages of the training process.
\label{fig8}}
\end{figure*}

\subsection{Classification accuracy}

The classification accuracies of the mentioned three tasks are shown in Fig. \ref{fig7} and Fig. \ref{fig8}. For reasons of the different distributional characteristics and classification challenges of these three cases, we designed different evaluation metrics for each algorithm. MNIST and UCI-HAR are not inherently unbalanced, but each class is not evenly distributed in the time domain. Therefore we introduced $N_{latest}$ mode into these two tasks to enhance data heterogeneity in each round and among different clients but still evaluated them with overall accuracy (Acc.). As for the case study of The Ford Challenge, due to that only about $16\%$ of the training data are of the positive label, we just set the training process as the standard mode but quantitatively compared our method with existing ones in the classification accuracy of minority classes (Acc.M). We also offered this setting on downsampled MNIST datasets for reference. Since the features of FedHealth were not specific to the MNIST dataset, the performance of FedHealth was not referenced in the MNIST case study. The experiment results are shown in Fig. \ref{fig7} for the Ford Challenge and Fig. \ref{fig8} for MNIST and UCI-HAR, showing that our method could overcome the negative impact of class imbalance issues and achieve better performance and outperform current mainstream, regardless of whether the imbalance issue occurs at the global level or between clients.

\begin{figure}[!h]
\centering{\includegraphics[width=0.8\linewidth]{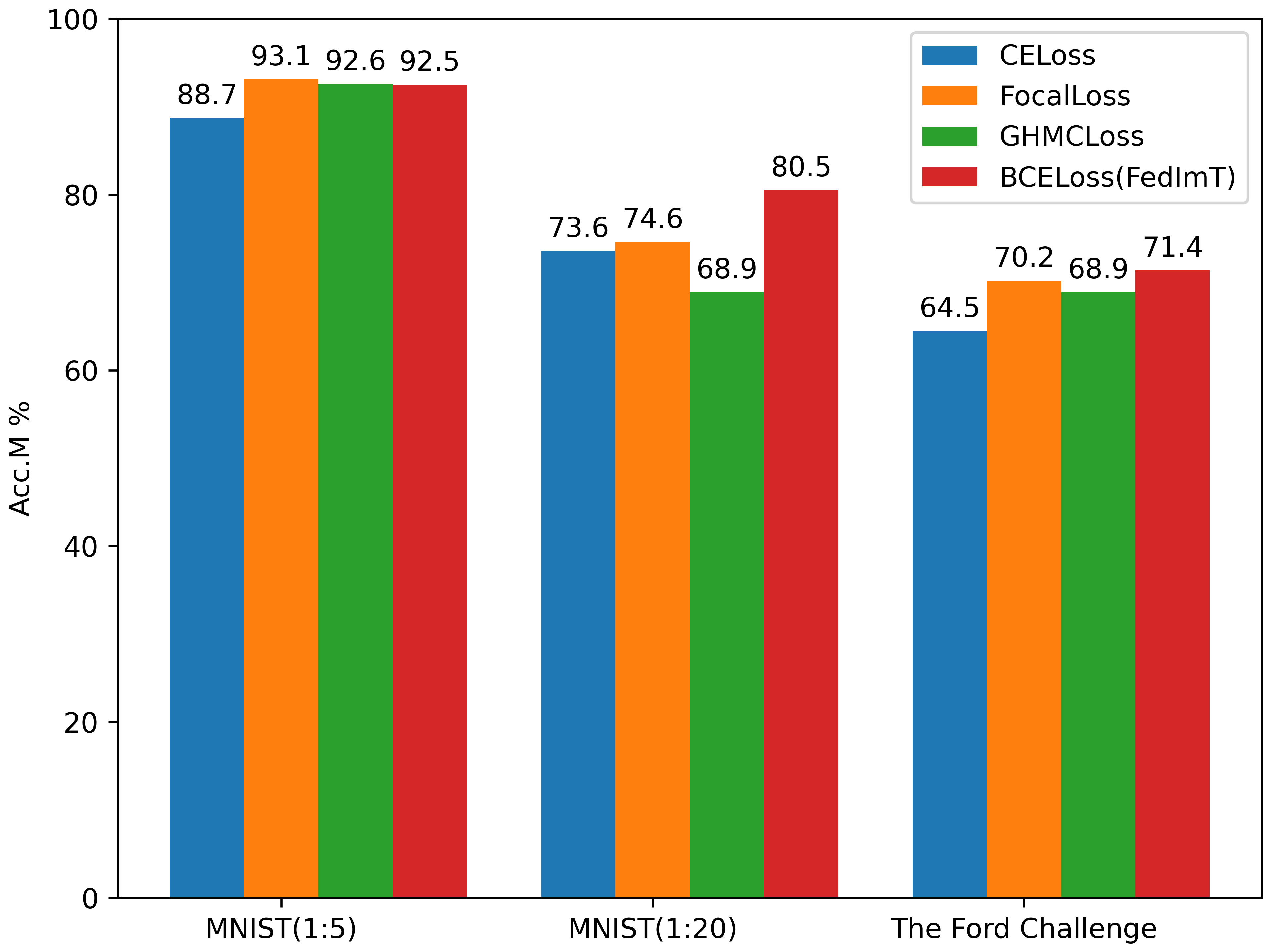}}
\caption{Performance comparison between our method and existing loss method in federate learning, over the defined case studies.
\label{fig6}}
\end{figure}

Moreover, considering that our approach improved classification performance by continuously rebalancing the loss function in the whole training process, we also compared our method with other kinds of loss functions that are designed for class imbalance in downsampled MNIST and drowsiness-detecting scenarios. We chose Focal Loss \cite{b32} and GHMC Loss \cite{b35} as representatives, together with the standard Cross-Entropy loss function without balance operation, to participate in the comparison. Notice that the Focal Loss and GHMC Loss here would not be offered the ratio observation result in the training process. Fig. \ref{fig6} demonstrates that in our defined cases and settings, our method still outperformed the other three in most scenarios. This shows the broader potential of the idea of optimizing the loss computation continuously during federated training.

\subsection{Detail analysis}

We provided comprehensive comparative experiments shown in Fig. \ref{fig7}, Fig. \ref{fig8}, and Fig. \ref{fig6}. Our algorithm can achieve similar performance with several state-of-the-art approaches. In addition to the improvement in accuracy, another significant benefit of our method is that it offers an always-on monitoring mechanism for training FL tasks, letting administrators be able to solve class imbalance issues at a very early stage to prevent negative impacts from happening.

Throughout all the experiment results, we observed that FedImT converged significantly slower than other algorithms in the early training process. We suggested that this was caused by the mismatch between the ratio observations in each round, thus activating the dropping mechanism of our method. This phenomenon decreased as the global round progressed. The results showed that forming a stable observation can significantly benefit the inferring performance.

Although the algorithmic complexity of the data composition estimating is quite high, due to the limited size of the auxiliary data, we did not observe noticeable extra processing duration and computation overhead in our experiments. an appropriate scale of auxiliary data can make the results of estimation more robust and precise.

\section{Conclusion}

In this paper, we proposed FedImT, an efficient imbalance-aware federated learning framework. FedImT was designed to meet the requirements of scenarios like medical intelligence or wearable healthcare, which have high privacy concerns and significant class imbalance issues. FedImT inherited the standard structure of FedAvg, where the server estimates the training data composition during each round of model aggregation, then autoregressions its estimation to form a stable and precise knowledge of the composition of training data. The loss computation would keep being re-balanced by such autoregressive estimation without the negative impact of class imbalance. The results demonstrated that FedImT achieves significant performance gains compared to representative baselines in terms of improved model accuracy and practicality.

\section*{REFERENCES}

\def\refname{\vadjust{\vspace*{-1em}}} %Please don't do this in a real paper.

\begin{IEEEbiography}[{\includegraphics[width=1in,height=1.25in,clip,keepaspectratio]{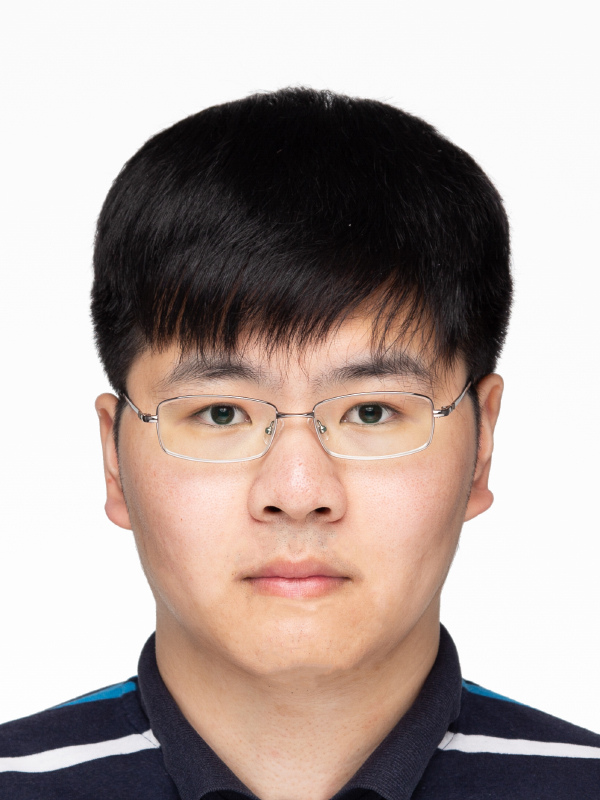}}]{Wenhao Yan } was born in Yangzhou, China. He received the B.S. degree in the College of Information Engineering from Yangzhou University, China. Then he has been an employee of Shenzhen ZTE NetView Technology as a senior data scientist in the data science and AI department since 2018. He was a research student with the Emerging Networks and Systems Laboratory (ENeS) and received the M.S. degree in the year 2023 at Muroran Institute of Technology, Japan. His research interests include federated learning, IoT, and distributed systems.
\end{IEEEbiography}

\begin{IEEEbiography}[{\includegraphics[width=1in,height=1.25in,clip,keepaspectratio]{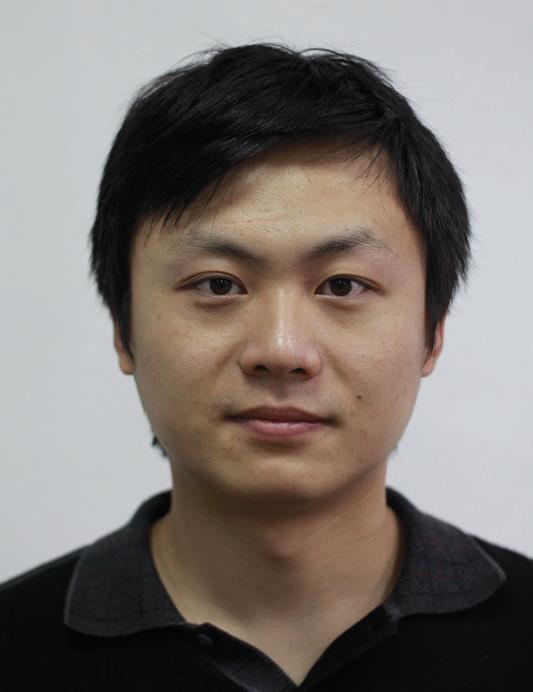}}]
{He Li } received the B.S., M.S. degrees in Computer Science and Engineering from Huazhong University of Science and Technology in 2007 and 2009, respectively, and Ph.D. degree in Computer Science and Engineering from The University of Aizu in 2015. He is currently an Associate Professor with Department of Sciences and Informatics, Muroran Institute of Technology, Japan. In 2018, he is selected as a Ministry of Education, Culture, Sports, Science and Technology (MEXT) Excellent Young Researcher. His research interests include IoT, edge computing, cloud computing and software defined networking. He has received the best journal paper awards from IEEE ComSoc APB and IEEE CSIM, and best paper awards from ICPADS 2019 and IEEE VTC2016-Fall. Dr. Li serves as an Associate Editor for Human-centric Computing and Information Sciences (HCIS), as well as Guest Associate Editors for Security and Communication Networks, Environments, and IEICE Transactions on Information and Systems. He is the recipient of 2019 IEEE TCSC Award for Excellence (Early Career Researcher) and 2016 IEEE TCSC Award for Excellence (Outstanding Ph.D Thesis).
\end{IEEEbiography}

\begin{IEEEbiography}[{\includegraphics[width=1in,height=1.25in,clip,keepaspectratio]{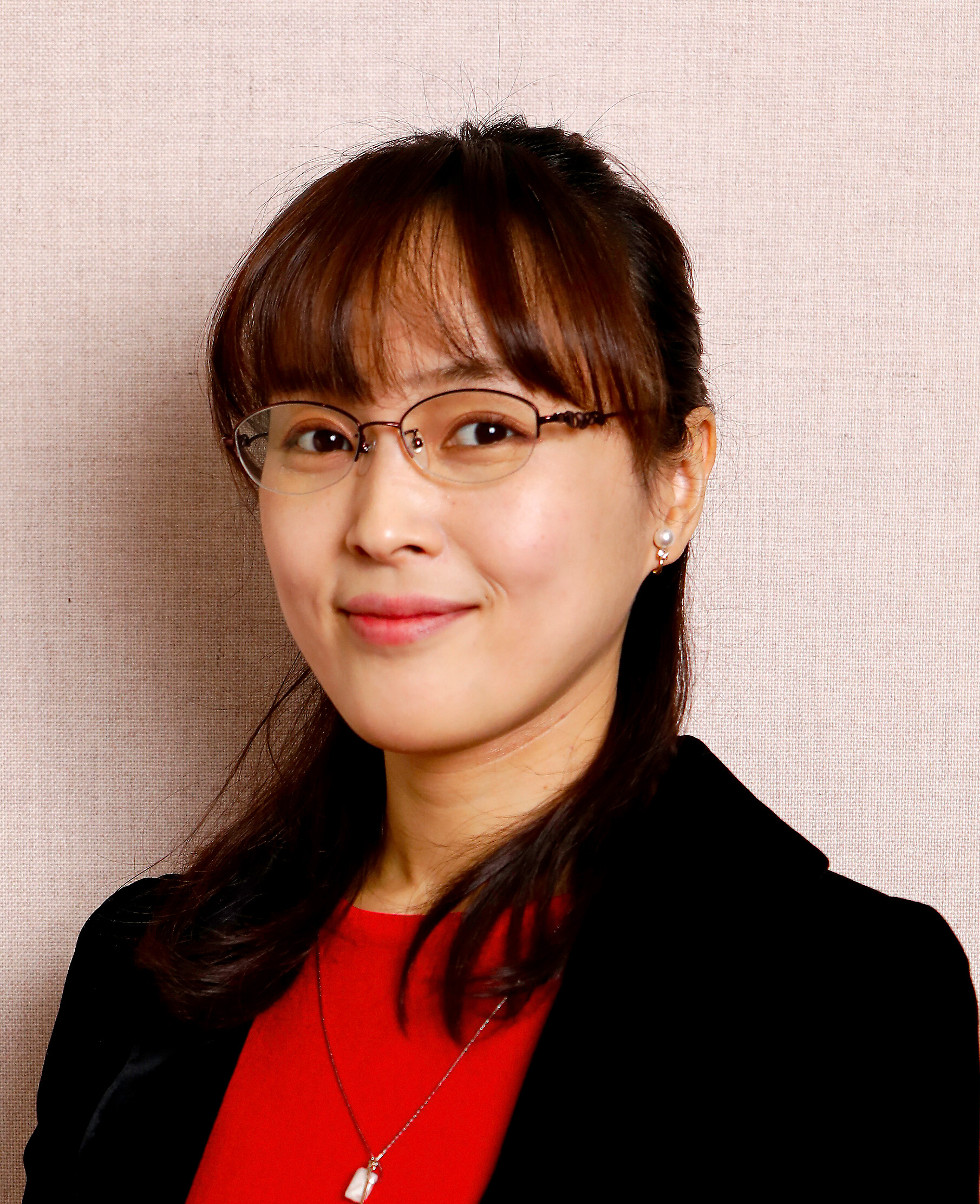}}]
{Kaoru
Ota }  was born in Aizu-Wakamatsu, Japan. She received M.S. degree in Computer Science from Oklahoma State University, the USA in 2008, B.S. and Ph.D. degrees in Computer Science and Engineering from The University of Aizu, Japan in 2006, 2012, respectively. Kaoru is a Professor and Ministry of Education, Culture, Sports, Science and Technology (MEXT) Excellent Young Researcher with the Department of Sciences and Informatics. She is also the founding Director of Center for Computer Science (CCS) at Muroran Institute of Technology, Japan. From March 2010 to March 2011, she was a visiting scholar at the University of Waterloo, Canada. Also, she was a Japan Society of the Promotion of Science (JSPS) research fellow at Tohoku University, Japan from April 2012 to April 2013. Kaoru is the recipient of IEEE TCSC Early Career Award 2017, The 13th IEEE ComSoc Asia-Pacific Young Researcher Award 2018, 2020 N2Women: Rising Stars in Computer Networking and Communications, 2020 KDDI Foundation Encouragement Award, and 2021 IEEE Sapporo Young Professionals Best Researcher Award, The Young Scientists’ Award from MEXT in 2023. She is Clarivate Analytics 2019, 2021, 2022 Highly Cited Researcher (Web of Science) and is selected as JST-PRESTO researcher in 2021, Fellow of EAJ in 2022.
\end{IEEEbiography}

\begin{IEEEbiography}[{\includegraphics[width=1in,height=1.25in,clip,keepaspectratio]{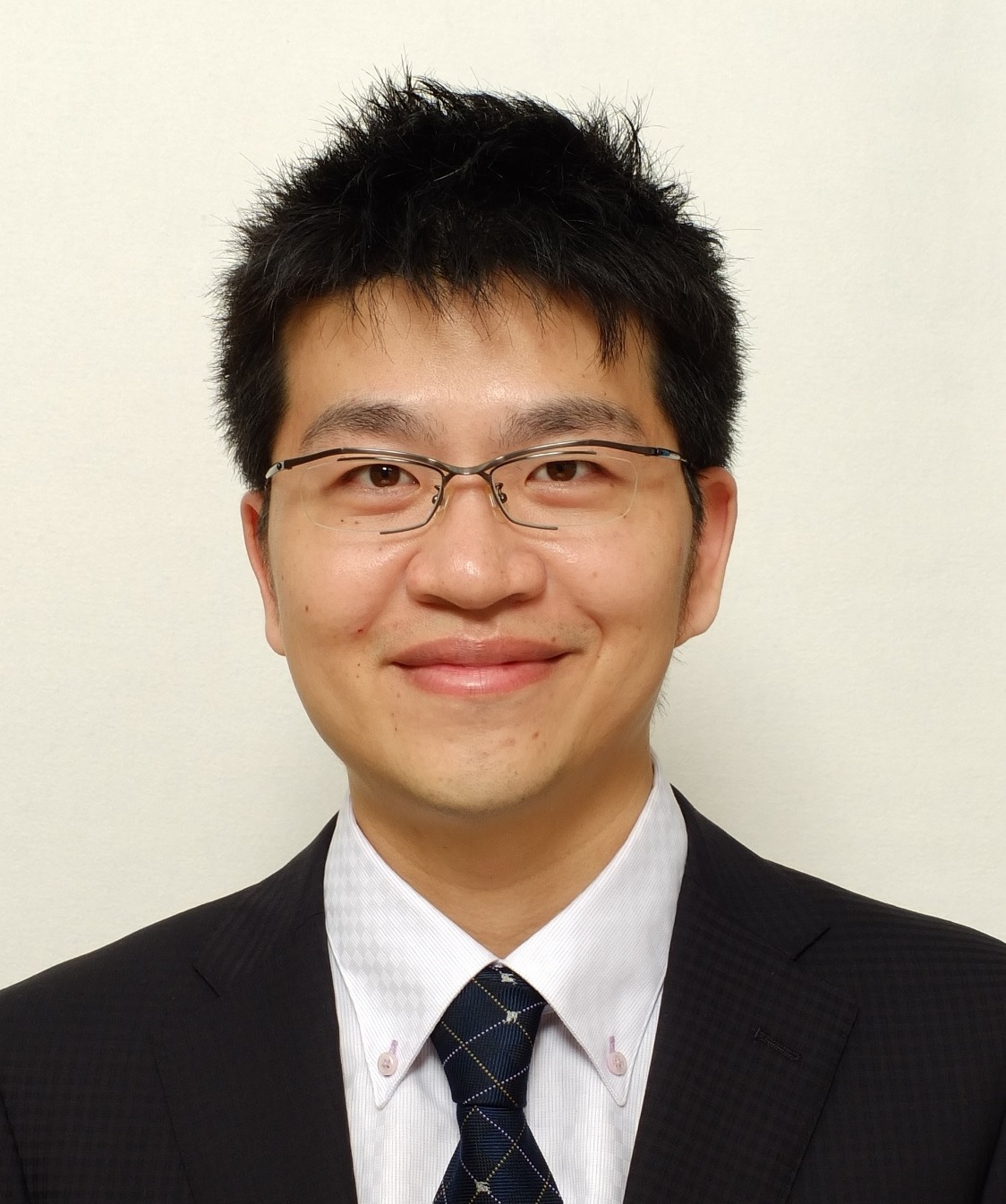}}]
{Mianxiong Dong } received B.S., M.S. and Ph.D. in Computer Science and Engineering from The University of Aizu, Japan. He is the Vice President and Professor of Muroran Institute of Technology, Japan. He was a JSPS Research Fellow with School of Computer Science and Engineering, The University of Aizu, Japan and was a visiting scholar with BBCR group at the University of Waterloo, Canada supported by JSPS Excellent Young Researcher Overseas Visit Program from April 2010 to August 2011. Dr. Dong was selected as a Foreigner Research Fellow (a total of 3 recipients all over Japan) by NEC C\&C Foundation in 2011. He is the recipient of The 12th IEEE ComSoc Asia-Pacific Young Researcher Award 2017, Funai Research Award 2018, NISTEP Researcher 2018 (one of only 11 people in Japan) in recognition of significant contributions in science and technology, The Young Scientists’ Award from MEXT in 2021, SUEMATSU-Yasuharu Award from IEICE in 2021, IEEE TCSC Middle Career Award in 2021. He is Clarivate Analytics 2019, 2021, 2022 Highly Cited Researcher (Web of Science) and Foreign Fellow of EAJ.
\end{IEEEbiography}


\begin{thebibliography}{00}
\bibitem{b01} N. Abbas, Y. Zhang, A. Taherkordi and T. Skeie, "Mobile Edge Computing: A Survey," \textit{IEEE Internet Things J.,} vol. 5, no. 1, pp. 450-465, Feb. 2018.

\bibitem{b02} G. D’Aniello, R. Gravina, M. Gaeta and G. Fortino, "Situation-Aware Sensor-Based Wearable Computing Systems: A Reference Architecture-Driven Review," \textit{IEEE Sens. J.,} vol. 22, no. 14, pp. 13853-13863, July 2022.

\bibitem{b03} Q. Li, Y. Diao, Q. Chen and B. He, "Federated Learning on Non-IID Data Silos: An Experimental Study," in \textit{Proc. Int. Conf. Data Eng.,} Kuala Lumpur, Malaysia, 2022, pp. 965-978.

\bibitem{b04} P. P. Ray, D. Dash, K. Salah and N. Kumar, "Blockchain for IoT-Based Healthcare: Background, Consensus, Platforms, and Use Cases," \textit{EEE Syst J.,} vol. 15, no. 1, pp. 85-94, March 2021.

\bibitem{b05} H. Li, K. Ota, M. Dong and M. Guo, "Learning Human Activities through Wi-Fi Channel State Information with Multiple Access Points," \textit{IEEE Commun. Mag.,} vol. 56, no. 5, pp. 124-129, May 2018.

\bibitem{b06} X. Zhou, W. Liang, J. Ma, Z. Yan and K. I. -K. Wang, "2D Federated Learning for Personalized Human Activity Recognition in Cyber-Physical-Social Systems," \textit{IEEE Trans. Netw. Sci.,} vol. 9, no. 6, pp. 3934-3944, 1 Nov.-Dec. 2022.

\bibitem{b07} Y. Lu, Y. -M. Cheung and Y. Y. Tang, "Bayes Imbalance Impact Index: A Measure of Class Imbalanced Data Set for Classification Problem," \textit{IEEE Trans. Neural Netw. Learn.,} vol. 31, no. 9, pp. 3525-3539, Sept. 2020.

\bibitem{b08} B. McMahan, E. Moore, D. Ramage, S. Hampson, and B. A. Y Arcas, "Communication-efficient learning of deep networks from decentralized data," in \textit{Proc. Artif. Intell. Statist.,} 2017, pp. 1273-1282.

\bibitem{b09} C. Düsing and P. Cimiano, "On the Trade-off Between Benefit and Contribution for Clients in Federated Learning in Healthcare," in \textit{Proc. 21st IEEE Int. Conf. Mach. Learn. Appl.,} Nassau, Bahamas, 2022, pp. 1672-1678.

\bibitem{b10} J.P.Albrecht, "How the GDPR will change the world," \textit{Eur. Data Prot. L. Rev.,} vol. 2, no. 3, pp. 287-289, 2016.

\bibitem{b11} X. Li, K. Huang, W. Yang, et al., "On the Convergence of FedAvg on Non-IID Data," in \textit{Proc. Int. Conf. Learn. Represent.,} Addis Ababa, Ethiopia, 2020.

\bibitem{b12} Y. Yang, S. Huang, W. Huang and X. Chang, "Privacy-Preserving Cost-Sensitive Learning," \textit{IEEE Trans. Neural Netw. Learn. Syst.,} vol. 32, no. 5, pp. 2105-2116, May 2021.

\bibitem{b13} S. Lu, Z. Gao, Q. Xu, C. Jiang, A. Zhang and X. Wang, "Class-Imbalance Privacy-Preserving Federated Learning for Decentralized Fault Diagnosis With Biometric Authentication," \textit{IEEE Trans. Ind. Inform.,} vol. 18, no. 12, pp. 9101-9111, Dec. 2022.

\bibitem{b14} N. D. Thong Tran, C. K. Leung, E. W. R. Madill and P. T. Binh, "A Deep Learning Based Predictive Model for Healthcare Analytics," in \textit{Proc. IEEE Int. Conf. Healthc. Inform.,} Rochester, MN, USA, 2022, pp. 547-549.

\bibitem{b15} T. K., G. K. Rajini and D. Maji, "Cost-Effective, Disposable, Flexible, and Printable MWCNT-Based Wearable Sensor for Human Body Temperature Monitoring," \textit{IEEE Sens. J.,} vol. 22, no. 17, pp. 16756-16763, Sept. 2022.

\bibitem{b16} H. Masnadi-Shirazi and N. Vasconcelos, "Cost-Sensitive Boosting," \textit{IEEE Trans. Pattern Anal. Mach. Intell.,} vol. 33, no. 2, pp. 294-309, Feb. 2011.

\bibitem{b17} Y. Wang, G. Gui, H. Gacanin, B. Adebisi, H. Sari and F. Adachi, "Federated Learning for Automatic Modulation Classification Under Class Imbalance and Varying Noise Condition," \textit{IEEE Trans. Cogn. Commun. Netw.,} vol. 8, no. 1, pp. 86-96, March 2022.

\bibitem{b18} R. Diversi, R. Guidorzi and U. Soverini, "Kalman filtering in extended noise environments," \textit{IEEE Trans. Autom. Control,} vol. 50, no. 9, pp. 1396-1402, Sept. 2005.

\bibitem{b19} Q. Jin and H. Ochiai, "Decentralized P2P Federated Learning on Ad-hoc Like Networks with Non-IID Dataset," in \textit{Proc. 18th Int. Conf. Wirel. Mob. Comput. Netw. Commun.,} Thessaloniki, Greece, 2022, pp. 387-393.

\bibitem{b20} M. Duan et al., "Astraea: Self-Balancing Federated Learning for Improving Classification Accuracy of Mobile Deep Learning Applications," in \textit{Proc. IEEE Int. Conf. Comput. Des.,} Abu Dhabi, United Arab Emirates, 2019, pp. 246-254.

\bibitem{b21} Y. Tao and S. S. -T. Yau, "Outlier-Robust Iterative Extended Kalman Filtering," \textit{IEEE Signal Process. Lett.,} vol. 30, pp. 743-747, 2023.

\bibitem{b22} D. Y. Zhang, Z. Kou, D. Wang, "FedSens: A Federated Learning Approach for Smart Health Sensing with Class Imbalance in Resource Constrained Edge Computing," in \textit{Proc. IEEE INFOCOM,} Vancouver, BC, Canada, 2021, pp. 1-10.

\bibitem{b23} L. Wang, S. Xu, X. Wang, et al., "Addressing class imbalance in federated learning," in \textit{Proc. AAAI Conf. Artif. Intell.,} vol. 35, no. 11, pp. 10165-10173, May 2021.

\bibitem{b24} H. Li, K. Ota and M. Dong, "Learning IoT in Edge: Deep Learning for the Internet of Things with Edge Computing," \textit{IEEE Netw.,} vol. 32, no. 1, pp. 96-101, Jan.-Feb. 2018.

\bibitem{b25} Latha R and Vetrivelan P, "Blood Viscosity based Heart Disease Risk Prediction Model in Edge/Fog Computing," in \textit{Proc. 11th Int. Conf. Commun. Syst. Netw.,} Bengaluru, India, 2019, pp. 833-837.

\bibitem{b26} G. Lin et al., "Human Activity Recognition Using Smartphones With WiFi Signals," \textit{IEEE Trans. Hum.-Mach. Syst.,} vol. 53, no. 1, pp. 142-153, Feb. 2023,

\bibitem{b27} T. DeVries and G. W. Taylor, "Improved regularization of convolutional neural networks with cutout," 2017,  \textit{arXiv:1708.04552}.

\bibitem{b28} E. D. Cubuk, B. Zoph, D. Mane, et al., "Autoaugment: Learning augmentation policies from data," 2018, \textit{arXiv:1805.09501}.

\bibitem{b29} X. Sun and N. Ansari, "EdgeIoT: Mobile Edge Computing for the Internet of Things," \textit{IEEE Commun. Mag.,} vol. 54, no. 12, pp. 22-29, December 2016.

\bibitem{b30} Y. Chen, X. Qin, J. Wang, C. Yu and W. Gao, "FedHealth: A Federated Transfer Learning Framework for Wearable Healthcare," \textit{IEEE Intell. Syst.,} vol. 35, no. 4, pp. 83-93, 1 July-Aug. 2020.

\bibitem{b31} Y. LeCun, L. Bottou, Y. Bengio, and P. Haffner. "Gradient-based learning applied to document recognition," \textit{Proc. IEEE,} vol. 86, no. 11, pp. 2278-2324, Nov. 1998.

\bibitem{b32} T. -Y. Lin, P. Goyal, R. Girshick, K. He and P. Dollár, "Focal Loss for Dense Object Detection," \textit{IEEE Trans. Pattern Anal. Mach. Intell.,} vol. 42, no. 2, pp. 318-327, Feb. 2020.

\bibitem{b33} Y. Li and W. Cao, "An Extended Multilayer Perceptron Model Using Reduced Geometric Algebra," \textit{IEEE Access,} vol. 7, pp. 129815-129823, 2019.

\bibitem{b34} S. H. Khan, M. Hayat, M. Bennamoun, F. A. Sohel and R. Togneri, "Cost-Sensitive Learning of Deep Feature Representations From Imbalanced Data," \textit{IEEE Trans. Neural Netw. Learn. Syst.,} vol. 29, no. 8, pp. 3573-3587, Aug. 2018.

\bibitem{b35} B. Li, Y. Liu, and X. Wang, "Gradient harmonized single-stage detector," in \textit{Proc. AAAI Conf. Artif. Intell.,} 2019, vol. 33, no. 1, pp. 8577-8584.

\bibitem{b36} Kaggle, "Stay alert! the ford challenge," [Online]. Available: \underline{https://www.kaggle.com/c/stayalert}. Accessed on: Mar. 1, 2023.

\bibitem{b37} T. Li, et al., "Federated optimization in heterogeneous networks," in \textit{Proc. Machine learning and systems (MLSys),} 2020, vol. 2, pp. 429-450.

\bibitem{b38} Wang Jianyu, et al., "Tackling the objective inconsistency problem in heterogeneous federated optimization." in \textit{Proc. Adv. Neural Inf. Process.,} 2020, vol. 33, pp. 7611-7623.

\bibitem{b39} Reyes-Ortiz Jorge, et al., "Human Activity Recognition Using Smartphones," \textit{UCI Machine Learning Repository,} [Online]. Available: \underline{https://doi.org/10.24432/C54S4K}. Accessed on: Mar. 1, 2023

\bibitem{b40} Y. Cui, M. Jia, T. -Y. Lin, Y. Song and S. Belongie, "Class-Balanced Loss Based on Effective Number of Samples," in \textit{Proc. IEEE Comput. Soc. Conf. Comput. Vis. Pattern Recognit.,} Long Beach, CA, USA, 2019, pp. 9260-9269.

\bibitem{b41} A. Pandey, P. Tiwary, S. Kumar and S. K. Das, "Adaptive Mini-Batch Gradient-Ascent-Based Localization for Indoor IoT Networks Under Rayleigh Fading Conditions," \textit{IEEE Internet of Things Journal,} vol. 8, no. 13, pp. 10665-10677, July 2021,

\bibitem{b42} Q. Li et al., "A Survey on Federated Learning Systems: Vision, Hype and Reality for Data Privacy and Protection," \textit{IEEE Trans. Knowl. Data Eng.,} vol. 35, no. 4, pp. 3347-3366, April 2023.
\end{thebibliography}
\end{document}